\documentclass[runningheads]{llncs}

\usepackage{eccv}

\usepackage{eccvabbrv}

\usepackage{graphicx}
\usepackage{booktabs}

\usepackage[accsupp]{axessibility}  

\usepackage{orcidlink}

\usepackage{amsmath}
\usepackage{multirow}
\usepackage{textcomp, gensymb}
\usepackage{dsfont}
\usepackage{tabularx}
\usepackage{algorithm}
\usepackage{algpseudocode}
\usepackage{color}
\usepackage{colortbl}
\usepackage{amsfonts}
\newcommand{\OURMODEL}{{$\text{F}^4$Splat}}
\newcommand{\tec}{\text{ctx}}
\newcommand{\tet}{\text{tgt}}

\begin{document}

\title{\texorpdfstring{F\textsuperscript{4}Splat: Feed-Forward Predictive Densification for Feed-Forward 3D Gaussian Splatting}{F4Splat: Feed-Forward Predictive Densification for Feed-Forward 3D Gaussian Splatting}}

\titlerunning{F\textsuperscript{4}Splat: Feed-Forward Predictive Densification for Feed-Forward 3DGS}

\author{Injae Kim\inst{1}\orcidlink{0009-0003-5863-1390} \and
Chaehyeon Kim\inst{2}\orcidlink{0009-0002-6910-1267} \and
Minseong Bae\inst{1}\orcidlink{0009-0009-5793-5003} \and
Minseok Joo\inst{2}\orcidlink{0009-0002-6775-2445} \and
Hyunwoo J. Kim\inst{1}$^{\dagger}$\orcidlink{0000-0002-2181-9264}}

\authorrunning{I.~Kim et al.}

\institute{KAIST, Daejeon, Republic of Korea \and
Korea University, Seoul, Republic of Korea \\
\url{https://mlvlab.github.io/F4Splat}}

\maketitle
{\let\thefootnote\relax\footnotetext{$^{\dagger}$ Corresponding author.\vspace{-30pt}}}
\begin{abstract} 
Feed-forward 3D Gaussian Splatting methods enable single-pass reconstruction and real-time rendering. However, they typically adopt rigid pixel-to-Gaussian or voxel-to-Gaussian pipelines that uniformly allocate Gaussians, leading to redundant Gaussians across views. Moreover, they lack an effective mechanism to control the total number of Gaussians while maintaining reconstruction fidelity.
To address these limitations, we present \textbf{F\textsuperscript{4}Splat}, which performs \emph{\textbf{F}eed-\textbf{F}orward predictive densification for \textbf{F}eed-\textbf{F}orward 3D Gaussian \textbf{Splat}ting}, introducing a densification-score-guided allocation strategy that adaptively distributes Gaussians according to spatial complexity and multi-view overlap. Our model predicts per-region densification scores to estimate the required Gaussian density and allows explicit control over the final Gaussian budget without retraining. This spatially adaptive allocation reduces redundancy in simple regions and minimizes duplicate Gaussians across overlapping views, producing compact yet high-quality 3D representations.
Extensive experiments demonstrate that our model achieves superior novel-view synthesis performance compared to prior uncalibrated feed-forward methods, while using significantly fewer Gaussians. 
\keywords{Feed-Forward 3D Gaussian Splatting \and Compact 3DGS}
\end{abstract}
\section{Introduction}
\begin{figure}[t!]
        \centering
        \includegraphics[width=1\linewidth]{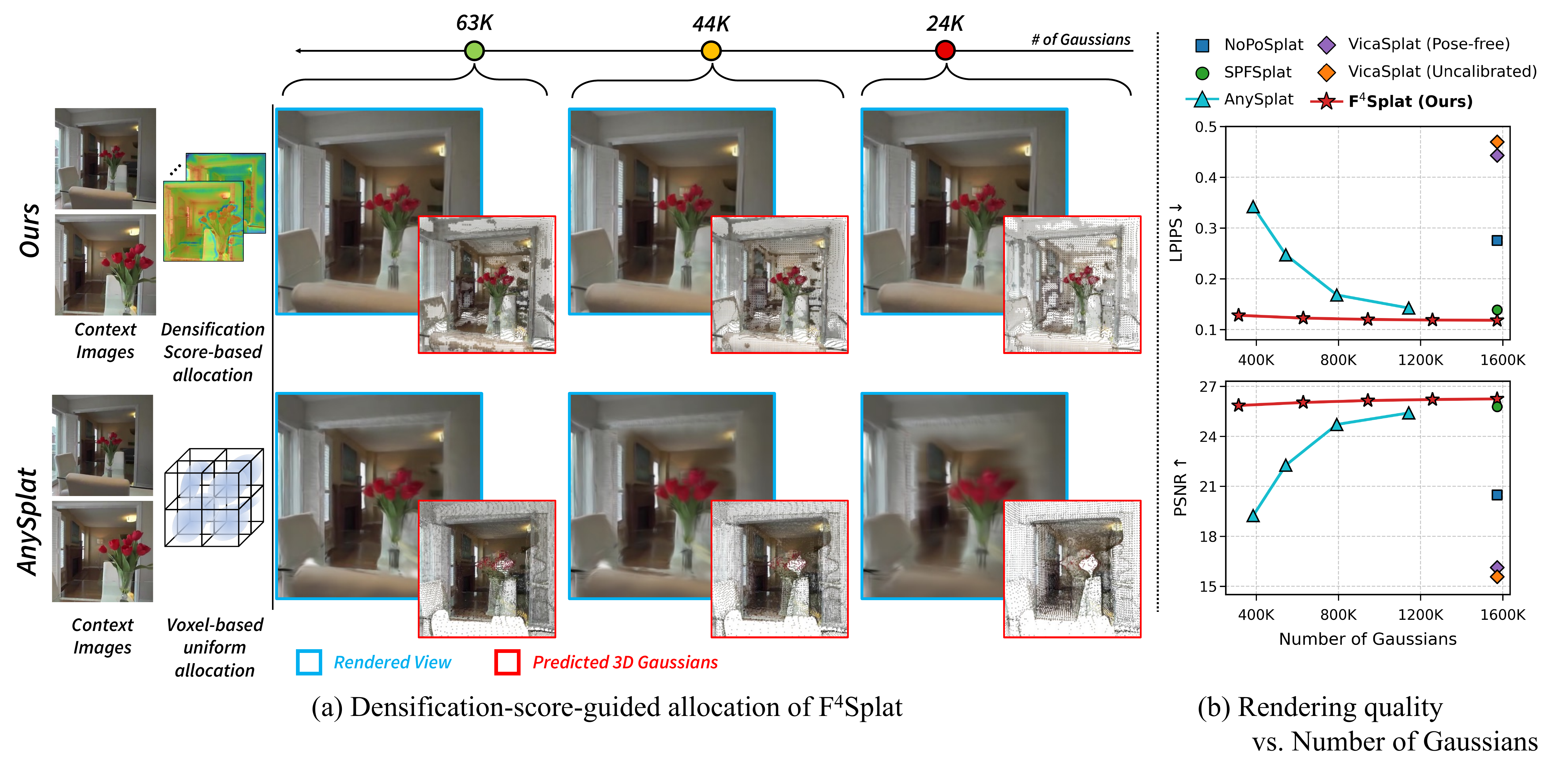}
         \setlength{\abovecaptionskip}{-10pt}
         \setlength{\belowcaptionskip}{-10pt}     
        \caption{\textbf{Comparison of Gaussian allocation under different target Gaussian budgets.} (a) Given the same target Gaussian budget, \OURMODEL\ allocates Gaussians non-uniformly using predicted densification scores, avoiding over-allocation in simple regions and concentrating primitives on fine-detail regions (e.g., flowers) to better preserve details even under a small budget, in contrast to voxel-based uniform allocation (AnySplat~\cite{jiang2025anysplat}). (b) Compared to pose-free and uncalibrated baselines, \OURMODEL\ achieves the best reconstruction fidelity (LPIPS \& PSNR), while largely preserving the rendering quality as the number of Gaussians is reduced.}
        \label{fig:teaser}
\end{figure}
In modern computer vision,  3D scene reconstruction using deep learning has become the de facto standard. In particular, 3D Gaussian Splatting (3DGS)~\cite{kerbl20233dgs} has emerged as a highly efficient alternative to existing methodologies~\cite{mescheder2019occupancynetworks, park2019deepsdf, fridovich2022plenoxels, muller2022instant}. 3DGS represents scenes using an explicit set of 3D Gaussian primitives, enabling high-fidelity 3D reconstruction and real-time novel-view rendering.
It incorporates adaptive density control (ADC), which periodically adds or removes Gaussians during optimization. These iterative updates assign a different number of Gaussians in each region, and through this adaptive assignment, the final 3DGS representation achieves high reconstruction fidelity with a relatively small number of Gaussians.
However, the conventional 3DGS framework still inherits key limitations shared by other optimization-based 3D reconstruction methods~\cite{mildenhall2021nerf, fridovich2022plenoxels, muller2022instant}. 
It requires costly per-scene iterative optimization and typically relies on densely captured input views with known camera parameters, which can be impractical in real-world scenarios. 
This has motivated feed-forward 3DGS methods~\cite{charatan2024pixelsplat, chen2024mvsplat, xu2025depthsplat, ye2024nopo, li2025vicasplat, huang2025spfsplat, jiang2025anysplat}, which are trained on large-scale datasets to build strong 3D priors. These frameworks can reconstruct a 3D scene from only a few input images through a single forward pass, preserving the real-time rendering capabilities of 3DGS while enabling generalization to unseen scenes. 
However, existing feed-forward 3DGS methods have a significant limitation in that they do not allocate Gaussians efficiently. This limitation stems from removing the iterative optimization process of conventional 3DGS, which also eliminates its periodic adaptive density control (ADC) that densifies Gaussians during training.
Most works~\cite{charatan2024pixelsplat, chen2024mvsplat, xu2025depthsplat, ye2024nopo, li2025vicasplat, zhang2025flare, huang2025spfsplat} adopt a pixel-to-Gaussian pipeline, which assigns Gaussians at the pixel level. This fixes the total number of Gaussians to the number of pixels in the entire image and prevents flexible adjustment of Gaussian positions, resulting in duplicated Gaussians across different views. AnySplat~\cite{jiang2025anysplat}, which employs a voxel-to-Gaussian pipeline, can adjust the number of Gaussians by changing the voxel size, but this typically requires training a new model. Moreover, because it allocates Gaussians uniformly in space (i.e., assigning one Gaussian per voxel), it struggles to produce a high-quality and compact Gaussian representation under a limited Gaussian budget.

To address inefficient Gaussian allocation, we introduce \textbf{\OURMODEL}, a feed-forward network that performs \emph{predictive densification} for 3D Gaussian Splatting from a set of uncalibrated images.
Our approach predicts densification decisions in a single forward pass, treating Gaussian densification as a learnable prediction problem within a unified feed-forward pipeline.
Specifically, the network estimates a \emph{densification score} that indicates whether additional Gaussians should be allocated to each region.
By estimating both spatial complexity and multi-view overlap, the predicted densification score avoids over-allocation in simple regions and prevents duplicate Gaussians in areas covered by overlapping input images.
This feed-forward densification strategy enables spatially adaptive allocation and yields compact Gaussian representations while maintaining competitive reconstruction fidelity. As illustrated in~\cref{fig:teaser}, this allows \OURMODEL\ to concentrate Gaussians on fine-detail regions while avoiding unnecessary allocation in simple regions, achieving higher rendering quality under the same Gaussian budget.
Through extensive experiments, \OURMODEL\ achieves on-par or superior novel-view synthesis quality while using significantly fewer Gaussians than prior uncalibrated feed-forward methods that rely solely on image inputs.

The contributions of our work can be summarized as:
\begin{itemize}
    \item \textbf{Gaussian-count controllable feed-forward 3DGS.} We propose \OURMODEL, a feed-forward framework that reconstructs 3D Gaussian Splatting representations from sparse, uncalibrated images while enabling explicit control over the final number of Gaussians through feed-forward predictive densification.

    \item \textbf{Densification-score-guided allocation for high fidelity under a limited budget.} We introduce a densification score that predicts where additional Gaussians should be allocated, enabling spatially adaptive Gaussian allocation without iterative optimization and maintaining high reconstruction fidelity even under a limited Gaussian budget.

    \item \textbf{State-of-the-art performance in the uncalibrated setting.} Extensive experiments show that \OURMODEL\ achieves on-par or superior novel-view synthesis quality while using significantly fewer Gaussians than prior uncalibrated feed-forward methods that rely solely on image inputs.
\end{itemize}
\section{Related Work}
\vspace{-5pt}
\noindent{\textbf{3D Gaussian Splatting for Novel View Synthesis.}} 
NeRF~\cite{mildenhall2021nerf} established a dominant paradigm for neural scene representation and sparked extensive subsequent research~\cite{martin2021nerf, barron2021mip, barron2022mip, yu2021plenoctrees, muller2022instant, fridovich2022plenoxels, park2021nerfies, barron2023zip, kim2023upnerf, chen2022tensorf}, driving rapid progress in neural scene reconstruction. However, its per-ray volumetric rendering incurs high compute cost, motivating more efficient alternatives.
3DGS~\cite{kerbl20233dgs} mitigates this inefficiency by representing a scene with a set of 3D Gaussian primitives and rendering them through differentiable rasterization, enabling real-time rendering and faster optimization.
To achieve high fidelity with a compact representation, it further employs \textit{adaptive density control} (ADC), which periodically adds Gaussians in under-represented regions and prunes primitives with negligible contribution during iterative optimization.
This has inspired a line of work on more compact 3DGS representations, spanning both refinements of the ADC strategy~\cite{ye2024absgs, rota2024revising, zhang2024pixel, kim2024color, kheradmand20243d, cheng2024gaussianpro, zhang2024fregs, grubert2025improving, fang2024mini, mallick2024taming} 
and various compaction and pruning methods~\cite{lee2024compact, niedermayr2024compressed, papantonakis2024reducing, fan2024lightgaussian, girish2024eagles, chen2024hac, wang2024end, yang2024spectrally}.

Despite the advantages of 3DGS, it still has several practical limitations. It typically assumes dozens to hundreds of diverse input views for stable reconstruction, which can be impractical in real-world scenarios. This dense-view requirement has been partly addressed by recent studies on \textit{sparse-view} 3DGS~\cite{xiong2023sparsegs, zhang2024cor, zhu2024fsgs, li2024dngaussian, he2025see, kong2025generative}, which aim to reconstruct 3D scenes from only a few input images. 
Another major limitation is that 3DGS still requires iterative per-scene optimization, which remains a significant burden for practical deployment. To reduce this time-consuming optimization process, a variety of recent approaches~\cite{feng2025flashgs, zhao2024grendel, hollein20253dgslm, chen2025dashgaussian, wang2025grouptraining3dgs} have sought to accelerate the original 3DGS optimization pipeline through more efficient rasterization, parallelization, and improved optimization strategies. Among these approaches, \textit{feed-forward 3DGS} represents a particularly promising paradigm, as it amortizes iterative optimization into a single feed-forward pass, thereby enabling much faster 3D reconstruction.

\noindent{\textbf{Feed-Forward 3D Gaussian Splatting.}} Feed-forward 3DGS approaches~\cite{charatan2024pixelsplat, chen2024mvsplat, wang2024freesplat, chen2024pref3r, smart2024splatt3r, ye2024nopo, zhang2025flare, hong2024pf3plat, kang2025selfsplat, huang2025spfsplat, li2025vicasplat, jiang2025anysplat} have been proposed to alleviate the costly per-scene optimization of standard 3DGS. These methods are trained on large-scale datasets to learn strong priors, allowing them to predict 3D Gaussian representations in a single feed-forward pass without iterative optimization.
Consequently, they can reconstruct from sparse views while enabling real-time rendering and generalization to unseen scenes. Early generalizable feed-forward 3DGS methods~\cite{charatan2024pixelsplat, chen2024mvsplat, wang2024freesplat} typically assume calibrated multi-view inputs with known camera poses. Recent works~\cite{chen2024pref3r, smart2024splatt3r, ye2024nopo, zhang2025flare} relax this assumption by moving to pose-free settings. In addition, self-supervised pose-free approaches~\cite{hong2024pf3plat, kang2025selfsplat, huang2025spfsplat} further reduce reliance on pose annotations by learning from reconstruction consistency, with pose estimation integrated into the pipeline. More recently, uncalibrated formulations~\cite{li2025vicasplat, jiang2025anysplat} are enabling reconstruction without camera calibration.

Despite these advances in efficiency and robustness, existing feed-forward 3DGS methods largely rely on a \emph{uniform} output parameterization, in which a fixed number of Gaussians is allocated per pixel or spatial unit. As a result, the total Gaussian count becomes tightly coupled with the input resolution, rather than being \emph{adaptively} allocated according to scene complexity. This leads to redundant primitives in simple regions, while failing to sufficiently model geometrically complex regions, resulting in a suboptimal and non-compact representation under a limited Gaussian budget. In conventional optimization-based 3DGS, this issue has been mitigated through adaptive density control (ADC), which dynamically allocates Gaussians based on scene structure. However, such mechanisms rely on iterative per-scene optimization and are therefore not directly applicable to feed-forward pipelines. The recent feed-forward approach, AnySplat~\cite{jiang2025anysplat}, is able to control the Gaussian count via voxel granularity. However, the allocation remains spatially uniform, and adapting to different budgets typically requires retraining, limiting flexibility and compactness. In contrast, we introduce Gaussian-count controllable feed-forward 3DGS, which predicts a budget-aware densification score that enables non-uniform and spatially adaptive Gaussian allocation. This yields a more compact 3D representation under a controllable Gaussian budget.
\section{Method}
We propose \textbf{\OURMODEL}, a feed-forward network that generates 3D Gaussian primitives~\cite{kerbl20233dgs} from an image collection via feed-forward predictive densification.
Unlike prior feed-forward 3DGS methods that rely on uniform allocation, our method allows users to adjust the number of Gaussians on demand through spatially adaptive Gaussian allocation, making more effective use of the available Gaussian budget.
In this section, we formulate the problem in \cref{sec3.1:problem_formulation}. Next, \cref{sec3.2:SADA} presents the overall framework, and \cref{sec3.3:training_pipeline} details the training pipeline.

\subsection{Problem Formulation} 
\label{sec3.1:problem_formulation}
Given $N_\tec$ input context images $\{\mathbf{I}^\tec_i\}^{N_\tec}_{i=1}$, where $\mathbf{I}^\tec_i \in \mathbb{R}^{3\times H\times W}$, most prior feed-forward 3D Gaussian Splatting (3DGS) works~\cite{chen2024mvsplat, xu2025depthsplat, ye2024nopo, li2025vicasplat, zhang2025flare, huang2025spfsplat} uniformly allocate one Gaussian per pixel, resulting in a fixed number of Gaussians,
$N_\tec HW$, to represent the scene. In contrast, our goal is to develop a feed-forward network $\mathbf{F}_\theta$ that enables control over the number of Gaussians.
Specifically, $\mathbf{F}_\theta$ takes as input not only the context images but also a user-specified target Gaussian budget $\bar{N}_{\mathcal{G}}$, and predicts a set of 3D Gaussian primitives $\mathcal{G} = \{\mathbf{g}_g\}^{N_{\mathcal{G}}}_{g=1}$, where $N_{\mathcal{G}}$ does not exceed $\bar{N}_{\mathcal{G}}$, and camera parameters $\{\hat{\mathbf{P}}^\tec_i\}^{N_\tec}_{i=1}$:
\begin{equation}
    \left( \{\mathbf{g}_g\}_{g=1}^{N_{\mathcal{G}}},\ \{ \hat{\mathbf{P}}^\tec_i\}^{N_\tec}_{i=1} \right) = \mathbf{F}_\theta\left(\{\mathbf{I}^\tec_i\}^{N_\tec}_{i=1},\ 
    \bar{N}_{\mathcal{G}} \right).
\end{equation}
Each Gaussian primitive ${\mathbf{g}_g} \in \mathbb{R}^{d_\mathbf{g}}$ is parameterized by center $\boldsymbol{\mu}_g\in\mathbb{R}^3$, opacity ${\sigma}_g \in \mathbb{R}$, rotation in quaternion ${\mathbf{q}}_g \in \mathbb{R}^4$, scale ${\mathbf{s}}_g \in \mathbb{R}^3$, and spherical harmonics (SH) ${\mathbf{h}}_g \in \mathbb{R}^\nu$.
Each camera parameter tuple is denoted as $\hat{\mathbf{P}}^\tec_i = (\hat{\mathbf{K}}^\tec_i, \hat{\mathbf{T}}^\tec_i)$,
where $\hat{\mathbf{K}}^\tec_i \in \mathbb{R}^{3\times3}$ is the intrinsic matrix and $\hat{\mathbf{T}}^\tec_i \in \mathbb{R}^{4\times4}$ is the camera-to-world pose.
We use $\hat{f}^\tec_i$ to denote the focal length in $\hat{\mathbf{K}}^\tec_i$. 
Here, $\hat{(\cdot)}$ denotes a predicted value.

It is not enough to merely control the number of Gaussians; it is also important to use them efficiently to generate a high-quality scene representation.
To this end, Gaussians should be allocated non-uniformly across the scene according to local characteristics, assigning more capacity to geometrically or visually complex regions.
In the case of AnySplat~\cite{jiang2025anysplat}, the number of Gaussians can be adjusted by changing the voxel size.
However, due to the inherent limitation of uniformly assigning a single Gaussian primitive to each voxel, it represents the scene less faithfully even under the same Gaussian count.
Uniform-Gaussian-allocation methods~\cite{charatan2024pixelsplat, chen2024mvsplat, xu2025depthsplat, ye2024nopo, li2025vicasplat, zhang2025flare, huang2025spfsplat, jiang2025anysplat} ignore the fact that different regions require different Gaussian densities, yielding redundant Gaussians in simple regions while under-allocating complex ones.
Consequently, the Gaussian budget is not spent where it is most needed for faithful scene representation.
To address this, we introduce a spatially adaptive Gaussian allocation framework that allocates Gaussians more effectively within a given budget.
\begin{figure*}[t]
     \centering
     \includegraphics[trim=10 0 0 0,clip, width=1\linewidth]{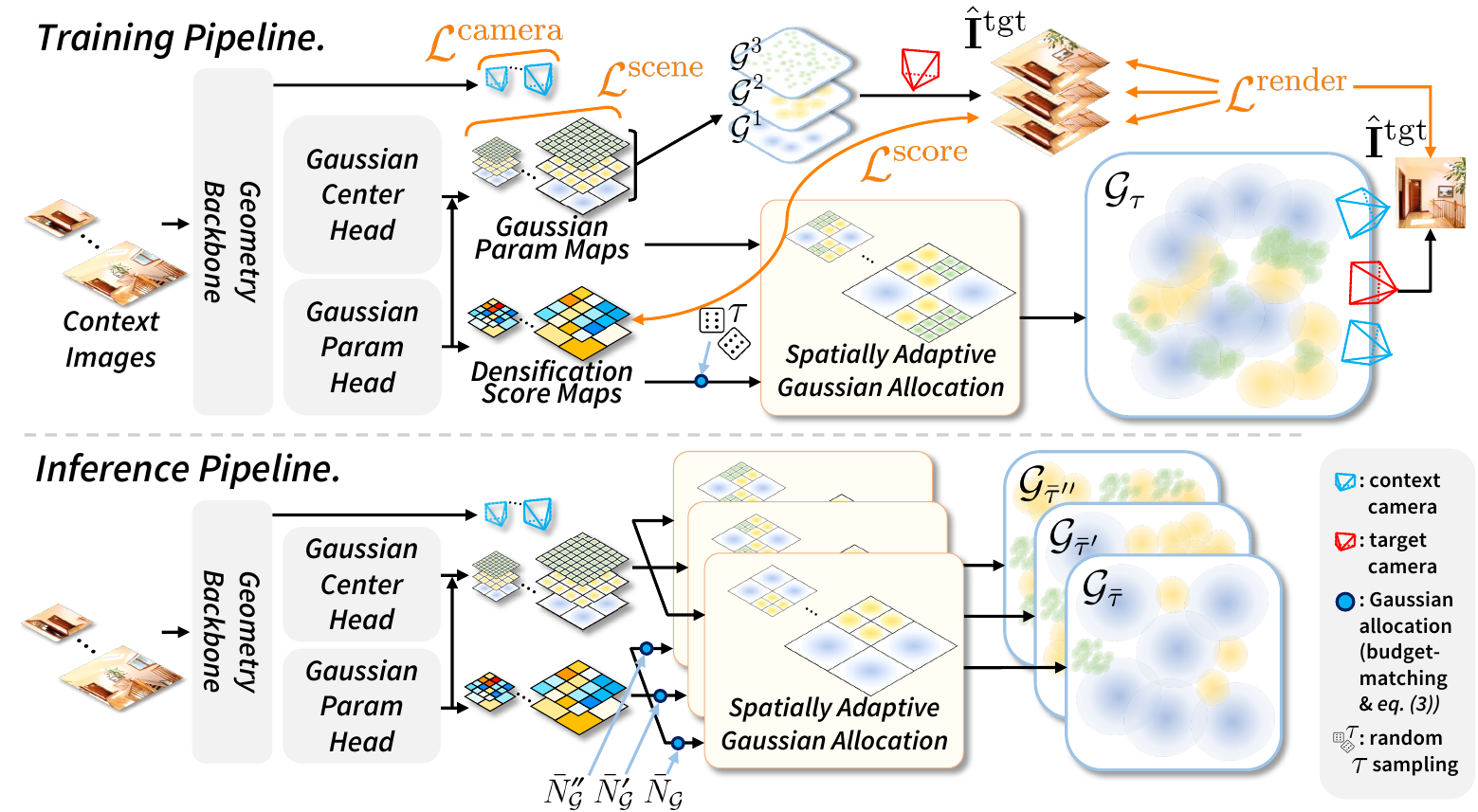}
     \setlength{\abovecaptionskip}{-5pt}
     \setlength{\belowcaptionskip}{-10pt}
     \caption{\textbf{Overview of \OURMODEL.}
    Given multi-view context images, the Geometry Backbone predicts camera parameters, multi-scale Gaussian parameter maps, and densification score maps. During training, camera and Gaussian parameters are jointly optimized with the camera loss $\mathcal{L}^{\text{camera}}$, rendering loss $\mathcal{L}^{\text{render}}$, and scene-scale regularization $\mathcal{L}^{\text{scene}}$. The predicted densification score maps are trained by score loss $\mathcal{L}^{\text{score}}$ derived from the backpropagation of rendering loss. The final representation $\mathcal{G}_{\tau}$, constructed using a randomly sampled threshold $\tau$, is also supervised by the rendering loss $\mathcal{L}^{\text{render}}$. 
    At inference time, given a target Gaussian budget $\bar{N}_{\mathcal{G}}$, the corresponding allocation threshold $\bar{\tau}$ is determined following the budget-allocation algorithm in Supplementary Sec.~S1. By varying $\bar{N}_{\mathcal{G}}$, the same model produces a family of compact, high-fidelity, Gaussian-count controllable representations $\mathcal{G}_{\bar{\tau}}$ without retraining.}
     \label{fig:main}
\end{figure*}
\subsection{Spatially Adaptive Gaussian Allocation}
\label{sec3.2:SADA}
As illustrated in~\cref{fig:main}, our framework consists of three parts: 
a \textit{Geometry Backbone} that encodes geometric information from a multi-view image set and predicts camera parameters; \textit{Gaussian Center Head} and \textit{Gaussian Parameter Head}, which predict multi-scale Gaussian parameter maps along with densification score maps; and \textit{Spatially Adaptive Gaussian Allocation} that effectively allocates the available Gaussian budget.

\noindent{\textbf{Geometry Backbone.}}
To encode geometric information from a given image set $\{\mathbf{I}^\tec_i\}_{i=1}^{N_\tec}$, we adopt a geometric backbone following the structure of VGGT~\cite{wang2025vggt}. 
Each input image $\mathbf{I}^\tec_i$ is processed by a pretrained DINOv2 encoder~\cite{oquab2023dinov2} to extract patch tokens. 
The resulting image tokens $t^\mathbf{I}_i$ are concatenated with learnable camera tokens $t^\mathbf{P}_i$ and register tokens $t^\text{r}_i$. 
The reference view has its own learnable camera and register tokens, while the remaining views share their corresponding tokens. 
The combined tokens $\{[t^\mathbf{I}_i; t^\mathbf{P}_i; t^\text{r}_i]\}_{i=1}^{N_\tec}$ are then passed through alternating frame-wise and global self-attention layers, resulting in the encoded tokens $\{[\tilde{t}^\mathbf{I}_i; \tilde{t}^\mathbf{P}_i; \tilde{t}^\text{r}_i]\}_{i=1}^{N_\tec}$. 
The encoded camera tokens $\{\tilde{t}^{\,\mathbf{P}}_i\}_{i=1}^{N_\tec}$ are passed through four additional self-attention layers, followed by a projection head to estimate camera parameters $\{\hat{\mathbf{P}}_i^\tec\}^{N_\tec}_{i=1}$. 

\noindent{\textbf{Multi-Scale Prediction.}}
To control the final number of Gaussians, multi-scale Gaussian parameter maps $\{\mathbf{G}^l_i\}_{l=1}^L$ and densification score maps $\{\hat{\mathbf{D}}^l_i\}_{l=1}^{L-1}$ are predicted from the encoded image tokens $\tilde{t}^{\,\mathbf{I}}_i$ encoded by the geometry backbone, where $\mathbf{G}^l_i\in \mathbb{R}^{d_\mathbf{g} \times H^l \times W^l}$ and $\hat{\mathbf{D}}^l_i\in \mathbb{R}^{H^l \times W^l}$.
We modify a DPT-based decoder~\cite{ranftl2021dpt} to introduce two parallel heads, a \textit{Gaussian Center Head} and a \textit{Gaussian Parameter Head}. 
Before the final two layers, the decoded feature maps are bilinearly interpolated to the target resolution $(H^l, W^l)$ at each level, and then the level-specific layers are applied. 
Each level-specific module consists of only two layers, enabling efficient multi-scale map prediction.
The Gaussian center head predicts the Gaussian centers, and the Gaussian parameter head predicts the remaining Gaussian primitives along with the densification score maps. In the Gaussian parameter head, an RGB shortcut~\cite{ye2024nopo} is utilized before the level-specific layers.
As the level $l$ increases, the spatial resolution doubles at each step, $(H^{l+1}, W^{l+1}) \!=\! (2H^{l}, 2W^{l})$. By exclusively selecting a scale level for each spatial region, we can control the final number of Gaussians $N_\mathcal{G}$. In the extreme case, selecting all regions from the coarsest level ($l\!=\!1$) yields $N_\tec H^1W^1$ Gaussians, while selecting all regions from the finest level ($l\!=\!L$) yields $N_\tec H^LW^L$ Gaussians. Therefore, $N_\mathcal{G}$ is bounded as:
\begin{equation}
    N_\tec H^1 W^1 \leq N_\mathcal{G} \leq N_\tec H^L W^L.
\end{equation}

\begin{figure*}[t]
     \centering
     \includegraphics[width=0.85\linewidth]{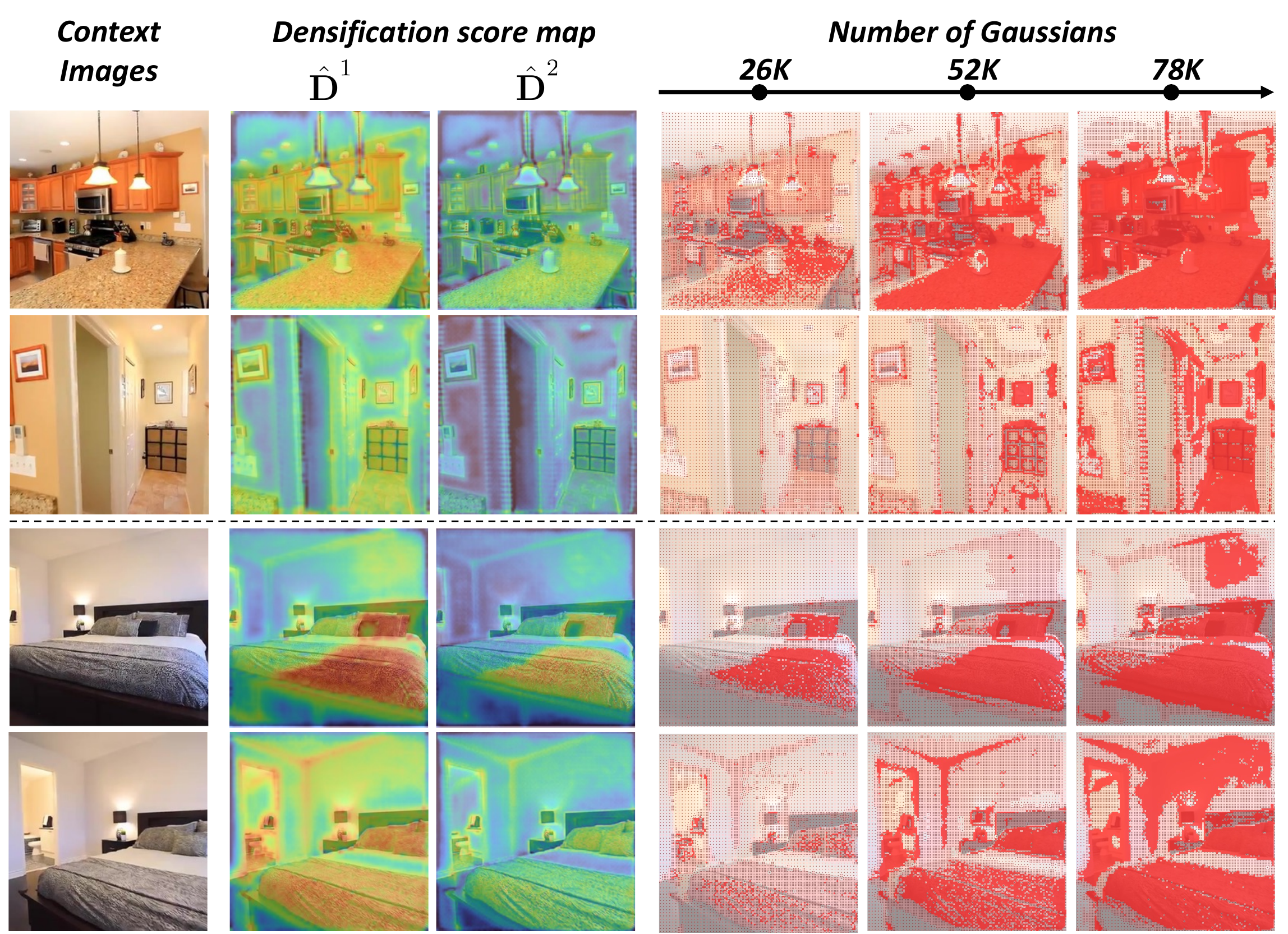}
     \caption{\textbf{Spatially adaptive Gaussian allocation.} Given two context images (left), the model predicts densification score maps that estimate where additional Gaussian density is required. 
     The \textcolor{red}{red} color indicates where more Gaussians are placed based on different fixed thresholds $\tau$. In the top example, we can clearly observe that more Gaussians are allocated in complex regions. The example below demonstrates redundancy-aware allocation across overlapping views, avoiding unnecessary allocations.}
     \label{fig:densification_score_analysis}
     \vspace{-10pt}
\end{figure*}
\noindent{\textbf{Spatially Adaptive Gaussian Allocation.}} 
Meanwhile, to represent a scene faithfully under a limited Gaussian budget, more Gaussians should be allocated to geometrically or photometrically complex regions. Additionally, redundant allocations to the same spatial locations across overlapping views should be minimized. If we can estimate how densely Gaussians should be in a given local space, we can allocate Gaussians more efficiently across the scene. 
To this end, we utilize a densification score map  $\hat{\mathbf{D}}^l_i \in \mathbb{R}^{H^l \times W^l}$, which indicates how densely Gaussians should be placed in each spatial region. 
As shown in Fig.~\ref{fig:densification_score_analysis}, thresholding the predicted densification scores yields spatially adaptive allocations, assigning more Gaussians to detail-rich regions while avoiding unnecessary allocation in simple or overlapping areas.
More details on the computation of the densification map are provided in the following~\cref{sec3.3:training_pipeline}.

Using the densification score maps, we determine the appropriate representation level for each region via a simple thresholding rule. Starting from the coarsest level ($l\!=\!1$), if the densification score is higher than a given threshold $\tau$, more Gaussians are allocated to that region from a higher-level Gaussian map. Ultimately, as illustrated in~\cref{fig:main}, Gaussians are selected such that allocations are non-overlapping across levels. The binary allocation masks $\mathcal{M}^l_{\tau,i} \in \{0,1\}^{H^l \times W^l}$, which indicate whether a particular location is allocated, can be computed as:
\vspace{-6pt}
\begin{equation}
\label{eq:feature_selection_mask}
    \mathcal{M}^l_{\tau,i} = 
    \begin{cases}
        \mathds{1}_{\{\hat{\mathbf{D}}^l_i < \tau\}}  & \text{if}\,\,\, l=1,  \\
        \mathds{1}_{\{\hat{\mathbf{D}}^l_i < \tau\}} \odot \left(\mathbf{1}-\sum_{k=1}^{l-1}\operatorname{Up}(\mathcal{M}^{l-k}_{\tau,i}; 2^k)\right)  & \text{if}\,\,\, 1<l<L,  \\
        \mathbf{1}-\sum_{k=1}^{l-1}\operatorname{Up}(\mathcal{M}^{l-k}_{\tau,i}; 2^k)  & \text{if}\,\,\, l = L,  \\
    \end{cases}
\end{equation}
where $\mathds{1}_{\{\cdot\}}$ is an indicator function that outputs $1$ when the condition is satisfied, $\mathbf{1}$ denotes a matrix of ones, $\odot$ is the element-wise product, and $\operatorname{Up}(\cdot;2^k)$ denotes the nearest-neighbor upsampling with a scaling factor of $2^k$. 
Using these masks, the final 3D Gaussian representation $\mathcal{G}_\tau\!=\!\{\mathbf{g}_g\}^{N_{\mathcal{G}_\tau}}_{g=1}$ is generated, where the total number of Gaussians $N_{\mathcal{G}_\tau}\!=\!\sum_{l,i}\|\mathcal{M}^l_{\tau,i}\|_1$.
Given a target Gaussian-count budget $\bar{N}_\mathcal{G}$, we can compute the minimum threshold $\bar{\tau}$ that satisfies the target budget by a simple budget-matching algorithm, since the Gaussians are exclusively selected across levels. The computed threshold $\bar{\tau}$ guarantees that the final number of Gaussians $N_{\mathcal{G}_{\bar{\tau}}}$ satisfies the following conditions:
{
\setlength{\abovedisplayskip}{3pt}
\setlength{\belowdisplayskip}{3pt}
\setlength{\abovedisplayshortskip}{2pt}
\setlength{\belowdisplayshortskip}{2pt}
\begin{equation}
\label{eq:budget_condition}
    0 \le \bar{N}_\mathcal{G} - N_{\mathcal{G}_{\bar{\tau}}}\!<4^{L-1}-1.
\end{equation}
}
The budget-matching algorithm is provided in the Supplementary Sec.~S1. 
\vspace{-3pt}
\subsection{Training Strategy}
\label{sec3.3:training_pipeline}
\noindent{\textbf{Feed-Forward Predictive Densification.}}
To allocate a limited number of Gaussians adaptively, the densification signal must satisfy two key properties. 
First, it should correlate with potential quality gain. It should allocate more Gaussians to under-represented regions and fewer to simple regions, so that representation quality tends to improve as the number of Gaussians increases.
Second, it must be available at inference time without requiring iterative optimization. A desirable densification score must be computable using only the input images. 
To satisfy these conditions, we draw inspiration from the adaptive density control (ADC) strategy of standard 3DGS works~\cite{kerbl20233dgs, ye2024absgs}, which iteratively optimizes 3D Gaussian primitives.

\begin{figure*}[t]
    \centering
    \includegraphics[width=1.0\textwidth]{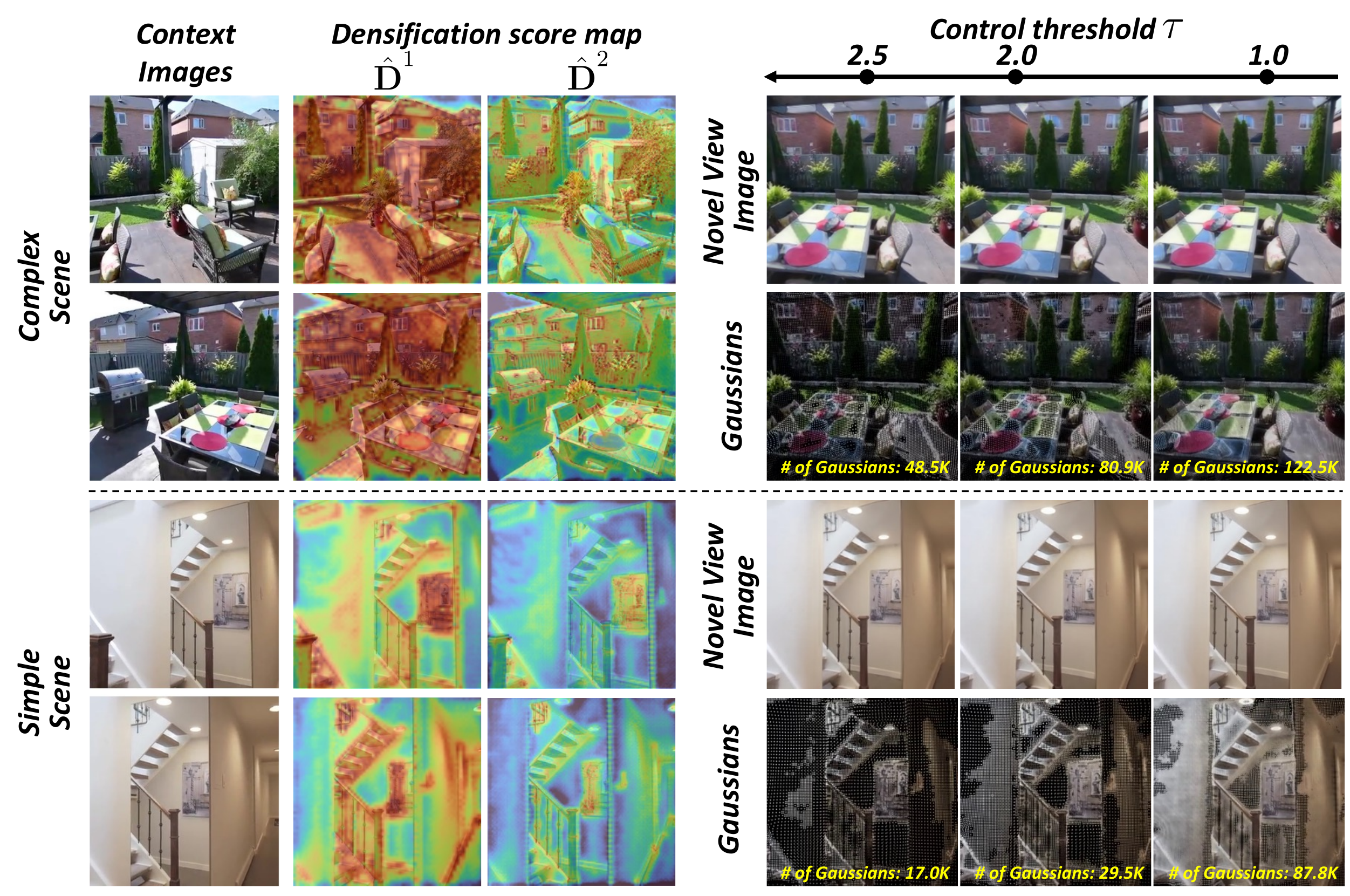}
    \vspace{-8pt}
        \caption{\textbf{Effect of the threshold $\tau$ across scene complexity.} For a fixed $\tau$, complex scenes yield higher densification scores and allocate more Gaussians, while simpler scenes allocate fewer.}
    \label{fig:threshold_analysis}
    \vspace{-15pt}
\end{figure*}

These previous works~\cite{kerbl20233dgs, ye2024absgs} periodically densify 3D Gaussians during training. In AbsGS~\cite{ye2024absgs}, for a set of predicted Gaussian primitives $\mathcal{G}=\{{\mathbf{g}_g}\}^{N_\mathcal{G}}_{g=1}$, whether to densify the Gaussian $\mathbf{g}_g$ is guided by the homodirectional target view-space positional gradient $\mathbf{v}_g$, which can be obtained by backpropagating the rendering loss. 
The rendering loss $\mathcal{L}^\text{render}$ is calculated between the predicted target image $\hat{\mathbf{I}}^\text{tgt}$ and the ground-truth target image $\mathbf{I}^\text{tgt}$.
With this loss, we can calculate the homodirectional view-space positional gradient $\mathbf{v}_g$:
\begin{equation}
    \mathbf{v}_g = \left( \,\sum^{m}_{j=1}\left|\frac{\partial \mathcal{L}_j^\text{render}}{\partial  \bar{\boldsymbol{\mu}}_{g,x}}\right|,\,\, \sum^{m}_{j=1}\left|\frac{\partial \mathcal{L}_j^\text{render}}{\partial \bar{\boldsymbol{\mu}}_{g,y}}\right|\,\, \right).
\end{equation}
Here $(\bar{\boldsymbol{\mu}}_{g,x}, \bar{\boldsymbol{\mu}}_{g,y})$ denotes the center of Gaussian $\mathbf{g}_g$ after projection onto the 2D image plane of the target view, $\mathcal{L}_j^\text{render}$ denotes the rendering loss computed by the $j$-th pixel of the image $\hat{\mathbf{I}}^\tet$, and $m$ is the total number of pixels that Gaussian $\mathbf{g}_g$ participates in rendering of $\hat{\mathbf{I}}^\tet$.
A large value of $\mathbf{v}_g$ indicates that the Gaussian $\mathbf{g}_g$ significantly affects the rendering loss, implying that the corresponding region is underrepresented. AbsGS~\cite{ye2024absgs} empirically shows that assigning more Gaussians based on the norm of this gradient improves the fidelity of the representation. Therefore, this gradient-based value is a suitable signal for our spatially adaptive Gaussian allocation pipeline. However, feed-forward 3DGS predicts 3D Gaussian primitives in a single forward pass without iterative scene optimization, and ground-truth images are not available at inference time. As a result, the gradient $\mathbf{v}_g$ cannot be utilized directly at inference time.

To overcome this limitation, we instead \textit{learn to predict} $\hat{d}_g$ from the gradient $\mathbf{v}_g$. Here, $\hat{d}_g$ denotes the predicted densification score associated with the spatial region occupied by the Gaussian $\mathbf{g}_g$. During the training process, using the 3D Gaussian representation $\mathcal{G}$ predicted by input context images $\{\mathbf{I}^\tec_i\}^{N_\tec}_{i=1}$, we can compute the rendering loss $\mathcal{L}^{\text{render}}(\hat{\mathbf{I}}^\tet,\mathbf{I}^\tet)$ and the homodirectional view-space positional gradient $\mathbf{v}_g$ for each Gaussian $\mathbf{g}_g\in\mathcal{G}$, as explained above. Based on the accumulated gradient $\mathbf{v}_g$, we define the supervision signal $d_g$ for the densification as a log-scaled value of the $\ell_2$ norm of the gradient: $d_g \!=\! \log\left(1+10^4\cdot \|\mathbf{v}_g\|_2\right)$. Then, the predicted densification score $\hat{d}_g$ for $\mathbf{g}_g$ is trained to match $d_g$ via the following $\ell_1$ loss:
\begin{equation}
    \mathcal{L}^{\text{score}}_\mathcal{G} = \mathbb{E}_{\mathbf{g}_g \in \mathcal{G}} \left[\|\hat{d}_g - d_g\|_1\right].
\end{equation}

This gradient-based supervision also implicitly captures cross-view redundancy: regions already well represented by overlapping context views produce lower densification signals, suppressing duplicate Gaussian allocation, as visualized in Fig.~\ref{fig:densification_score_analysis}. Furthermore, since the score is learned from gradient signals, it serves as an absolute and comparable criterion across different scenes, rather than a purely relative ranking within a single scene. Consequently, the threshold $\tau$ serves as a control knob for reconstruction fidelity, determining the level of detail preserved in the final representation. As shown in~\cref{fig:threshold_analysis}, under a fixed $\tau$, a complex scene generally produces higher densification scores, resulting in higher Gaussian allocation. In contrast, a simple scene yields lower scores and consequently fewer Gaussians. 

\noindent{\textbf{Training with Novel Views.}} Training the model by supervising the final 3D Gaussian representation using the input context views, rather than novel views, can lead to trivial reconstruction and potential overfitting, as the model may simply memorize the observed views instead of learning a representation that generalizes across viewpoints. To address this issue, unlike the previous feed-forward approach~\cite{jiang2025anysplat} that supervises reconstruction of input context views, we instead use a novel view as the target image during training. 
However, directly using the ground-truth camera parameters of the target view is problematic because the camera coordinate system predicted by the model may differ from the ground-truth coordinate frame. To resolve this mismatch, we align the ground-truth target camera into the predicted coordinate system before rendering the target view.
Specifically, let $\mathbf{T}^\tec_{n_1}$ and $\mathbf{T}^\tec_{n_2}$ denote the ground-truth poses of the two context views closest to the target view, and $\hat{\mathbf{T}}^\tec_{n_1}$ and $\hat{\mathbf{T}}^\tec_{n_2}$ denote their corresponding predicted poses. 
We estimate a similarity transformation matrix $\mathbf{A} \in \mathrm{Sim}(3)$
that maps poses from the ground-truth coordinate frame to the predicted one.
The context view $n_1$ closest to the target view is used as the anchor reference for rotation and translation alignment, while the scale is estimated from the relative translation between the two context views $n_1$ and $n_2$ closest to the target view.
The ground-truth target pose $\mathbf{T}^\tet$ is then transformed using the matrix $\mathbf{A}$ to obtain the aligned target pose, $\hat{\mathbf{T}}^\tet = \mathbf{A} \mathbf{T}^\tet$. For the intrinsic parameters $f$, we adjust the focal length of the target view using the ratio between the predicted and ground-truth focal lengths of the nearest context view, $\hat{f}^\tet = \hat{f}^\tec_{n_1}/ f^\tec_{n_1} \cdot f^\tet$. This simple yet carefully designed alignment procedure ensures that the aligned target view parameters $(\hat{\mathbf{T}}^\tet, \hat{\mathbf{K}}^\tet)$ are geometrically consistent with the coordinate system predicted by the model, enabling stable and reliable novel view training.

\noindent{\textbf{Overall Training Pipeline.}}
During training, we randomly sample a threshold $\tau$ and predict the corresponding Gaussian set $\mathcal{G}_\tau$, which is supervised using the rendering loss $\mathcal{L}^{\text{render}}$. In addition, we also apply the rendering loss to level-wise Gaussian representations $\mathcal{G}^l\in\mathbb{R}^{d_\mathbf{g}\times N_\tec H^l W^l}$, which are constructed using only the Gaussians predicted at each level $l$.
The densification score loss $\mathcal{L}^{\text{score}}$ is computed using the densification scores derived from the multi-level Gaussian representations $\{\mathcal{G}^l\}_{l=1}^{L-1}$. The loss for camera parameter estimation, $\mathcal{L}^{\text{camera}}$, is learned in the same way as VGGT~\cite{wang2025vggt}.
We further introduce a scene-scale regularization loss $\mathcal{L}^{\text{scene}}$, which normalizes the average distance of Gaussian centers from the origin to $1$, \emph{i.e.}, $\mathcal{L}^{\text{scene}} = \left|\frac{1}{|\mathcal{G}|}\sum_{\mathbf{g}_g\in\mathcal{G}} \|\boldsymbol{\mu}_g\|_2 - 1\right|$. This regularization is applied independently to the representation of each level. Without this regularization, training can become unstable. Finally, the total training objective is defined as the weighted sum of $\mathcal{L}^\text{render}$, $\mathcal{L}^\text{score}$, $\mathcal{L}^\text{camera}$, and $\mathcal{L}^\text{scene}$.

\section{Experiments}

\noindent{\textbf{Datasets.}}
We train \OURMODEL\ on the large-scale RealEstate10K (RE10K)~\cite{zhou2018re10k}, ACID~\cite{liu2021acid} datasets, and DL3DV~\cite{ling2024dl3dv}, following prior work~\cite{ye2024nopo} for train/test splits.
For two-view evaluation, we adopt the same test split as prior feed-forward methods~\cite{charatan2024pixelsplat, chen2024mvsplat, xu2025depthsplat, li2025vicasplat, ye2024nopo}. 
For multi-view evaluation, we use the scene categorization provided by NoPoSplat~\cite{ye2024nopo} and select input pairs with small overlap. Additional input views are sampled between the selected pair without duplication to match the target number of input views (8, 16, and 24 views).

\noindent{\textbf{Implementation Details.}} We use three levels of multi-resolution feature maps (\(L=3\)) in all experiments and choose the finest level to match the input image resolution, $(H^L, W^L)\!=\!(H,W)$. We initialize our model from pretrained VGGT weights. We use a learning rate of \(2\times10^{-4}\) for most modules, while freezing the patch embedding weights of the geometry backbone and training the remaining backbone parameters with a smaller learning rate of \(2\times10^{-5}\).
For multi-view training, we train on RE10K. At each iteration, each GPU independently samples the number of input views from $\{2,3,4,6,12,24\}$ and selects context images accordingly. We additionally sample the same number of target novel views for supervision. To keep the total number of images processed per iteration constant, we use a dynamic batch size that is inversely proportional to the number of context images. The multi-view model is trained for 15,000 iterations.
For two-view training, separate models are trained on RE10K and ACID, following the training configuration of NoPoSplat~\cite{ye2024nopo}. Each model is trained for 18,750 iterations with a total batch size of 128. We set the weight of MSE and LPIPS loss as 1 and 0.05 for $\mathcal{L}^\text{render}$. For the final total loss, we fix the weight of $\mathcal{L}^\text{render}$, $\mathcal{L}^\text{score}$, $\mathcal{L}^\text{camera}$, and $\mathcal{L}^\text{scene}$ as $1.0$, $10^{-4}$, $10.0$, and $10^{-2}$ respectively, in all experiments. All experiments are conducted on eight NVIDIA H200 GPUs, and each training run takes approximately 15 hours.

\noindent{\textbf{Baselines and Evaluation Metrics.}}
For multi-view evaluation, we compare \OURMODEL\ with recent feed-forward 3DGS methods under two settings: \textit{uncalibrated} methods that take only images as input without camera poses or intrinsics, and \textit{pose-free} methods that do not require camera poses but assume known intrinsics. 
For the uncalibrated setting, we compare against VicaSplat~\cite{li2025vicasplat} and AnySplat~\cite{jiang2025anysplat}. For fair comparison, we retrain AnySplat under the same training setup as ours, which requires approximately 27 hours on eight NVIDIA H200 GPUs. VicaSplat uses the officially released multi-view pretrained weights on the RE10K dataset. 
For the pose-free setting, we compare against NoPoSplat~\cite{ye2024nopo}, VicaSplat~\cite{li2025vicasplat}, and SPFSplat~\cite{huang2025spfsplat}, using their officially released pretrained models.
For two-view evaluation, we compare with pixelSplat~\cite{charatan2024pixelsplat}, MVSplat~\cite{chen2024mvsplat}, NoPoSplat~\cite{ye2024nopo}, VicaSplat~\cite{li2025vicasplat}, and SPFSplat~\cite{huang2025spfsplat}. We also compare baselines that are not a feed-forward 3D Gaussian splatting method, including pixelNeRF~\cite{yu2021pixelnerf}, DUSt3R~\cite{wang2024dust3r}, MASt3R~\cite{leroy2024mast3r}, and CoPoNeRF~\cite{hong2023coponerf}.
We evaluate the performance of novel view synthesis using three standard metrics: LPIPS, SSIM, and PSNR. We also report the number of final Gaussian primitives to enable joint comparison of rendering quality and Gaussian-count efficiency.
\vspace{-10pt}

\begin{table*}[t]
\centering
\setlength{\tabcolsep}{0.09cm}
\renewcommand{\arraystretch}{0.95}
\caption{\textbf{Novel view synthesis performance on RE10K~\cite{zhou2018re10k} under different numbers of input views.}
We report LPIPS/SSIM/PSNR for 8, 16, and 24 input views.
Best and second-best results are highlighted in \textbf{bold} and \underline{underlined}, respectively. $\tau^+$ and $\tau^-$ denote the high- and low-threshold variants, respectively.}
\label{tab:re10k_multiviews}
\vspace{-6pt}

\resizebox{\textwidth}{!}{
\begin{tabular}{llcccc cccc cccc}
\toprule
& \multirow{2}{*}{\textbf{Method}}
& \multicolumn{4}{c}{\textbf{8 views}}
& \multicolumn{4}{c}{\textbf{16 views}}
& \multicolumn{4}{c}{\textbf{24 views}} \\
\cmidrule(lr){3-6} \cmidrule(lr){7-10} \cmidrule(lr){11-14}
& 
& \#GS$\downarrow$ & LPIPS$\downarrow$ & SSIM$\uparrow$ & PSNR$\uparrow$
& \#GS$\downarrow$ & LPIPS$\downarrow$ & SSIM$\uparrow$ & PSNR$\uparrow$
& \#GS$\downarrow$ & LPIPS$\downarrow$ & SSIM$\uparrow$ & PSNR$\uparrow$ \\
\midrule

\multirow{3}{*}{\shortstack[l]{\emph{Pose-Free}}}
& NoPoSplat
& 524K & 0.213 & 0.756 & 22.31
& 1049K & 0.252 & 0.713 & 21.11
& 1573K & 0.275 & 0.691 & 20.49 \\

& VicaSplat
& 524K & 0.241 & 0.713 & 21.18 
& 1049K & 0.384 & 0.596 & 17.56
& 1573K & 0.443 & 0.546 & 16.13 \\

& SPFSplat
& 524K & 0.142 & 0.849 & 25.66
& 1049K & 0.137 & 0.855 & 25.88
& 1573K & 0.139 & 0.853 & 25.78 \\

\midrule

\multirow{4}{*}{\shortstack[l]{\emph{Uncalibrated}}}
& VicaSplat
& 524K & 0.258 & 0.686 & 20.77
& 1049K & 0.417 & 0.556 & 16.78
& 1573K & 0.470 & 0.517 & 15.58 \\

& AnySplat
& \underline{447K} & 0.167 & 0.819 & 24.07
& \underline{820K} & 0.148 & 0.842 & 25.10
& \underline{1142K} & 0.143 & 0.849 & 25.40 \\

& \textbf{\OURMODEL$_{\tau^+}$}
& \textbf{105K} & \underline{0.140} & \underline{0.851} & \underline{25.35}
& \textbf{210K} & \underline{0.126} & \underline{0.866} & \underline{25.91}
& \textbf{315K} & \underline{0.122} & \underline{0.870} & \underline{26.01} \\

& \textbf{\OURMODEL$_{\tau^-}$}
& \underline{447K} & \textbf{0.128} & \textbf{0.863} & \textbf{25.73}
& \underline{820K} & \textbf{0.116} & \textbf{0.875} & \textbf{26.28} 
& \underline{1142K} & \textbf{0.113} & \textbf{0.879} & \textbf{26.43} \\
\bottomrule
\end{tabular}
}
\vspace{-5pt}
\end{table*}

\begin{table*}[t]
\centering
\setlength{\tabcolsep}{0.09cm}
\renewcommand{\arraystretch}{0.95}
\caption{\textbf{Generalization to unseen datasets.} Models are trained on RE10K~\cite{zhou2018re10k} and evaluated on the unseen ACID dataset~\cite{liu2021acid}. We compare reconstruction quality under different numbers of input views (8, 16, and 24).}
\label{tab:crossdata_acid_nview}
\vspace{-6pt}

\resizebox{\textwidth}{!}{
\begin{tabular}{llcccc cccc cccc}
\toprule
& \multirow{2}{*}{\textbf{Method}}
& \multicolumn{4}{c}{\textbf{8 views}}
& \multicolumn{4}{c}{\textbf{16 views}}
& \multicolumn{4}{c}{\textbf{24 views}} \\
\cmidrule(lr){3-6} \cmidrule(lr){7-10} \cmidrule(lr){11-14}
& 
& \#GS$\downarrow$ & LPIPS$\downarrow$ & SSIM$\uparrow$ & PSNR$\uparrow$
& \#GS$\downarrow$ & LPIPS$\downarrow$ & SSIM$\uparrow$ & PSNR$\uparrow$
& \#GS$\downarrow$ & LPIPS$\downarrow$ & SSIM$\uparrow$ & PSNR$\uparrow$ \\
\midrule

\multirow{3}{*}{\shortstack[l]{\emph{Pose-Free}}}
& NoPoSplat
& 524K & 0.268 & 0.672 & 22.84
& 1049K & 0.294 & 0.644 & 22.14
& 1573K & 0.307 & 0.630 & 21.78 \\

& VicaSplat
& 524K & 0.271 & 0.656 & 22.57
& 1049K & 0.327 & 0.609 & 21.20
& 1573K & 0.356 & 0.581 & 20.37 \\

& SPFSplat
& 524K & 0.215 & 0.736 & 24.89
& 1049K & 0.214 & 0.743 & 25.07
& 1573K & 0.218 & 0.744 & 25.00 \\

\midrule

\multirow{4}{*}{\shortstack[l]{\emph{Uncalibrated}}}
& VicaSplat
& 524K & 0.315 & 0.588 & 21.31
& 1049K & 0.377 & 0.547 & 19.82
& 1573K & 0.399 & 0.529 & 19.22 \\

& AnySplat
& \underline{481K} & 0.248 & 0.696 & 23.30 
& \underline{906K} & 0.236 & 0.720 & 23.88
& \underline{1289K} & 0.234 & 0.727 & 24.04 \\

& \textbf{\OURMODEL$_{\tau^+}$}
& \textbf{52K} & \underline{0.230} & \underline{0.726} & \underline{24.55}
& \textbf{105K} & \underline{0.215} & \underline{0.745} & \underline{25.02}
& \textbf{315K} & \underline{0.195} & \underline{0.768} & \underline{25.51} \\

& \textbf{\OURMODEL$_{\tau^-}$}
& \underline{481K} & \textbf{0.194} & \textbf{0.759} & \textbf{25.16}
& \underline{906K} & \textbf{0.184} & \textbf{0.775} & \textbf{25.56}
& \underline{1289K} & \textbf{0.181} & \textbf{0.780} & \textbf{25.70} \\
\bottomrule
\end{tabular}
}
\vspace{-15pt}
\end{table*}
\subsection{Experimental Results}
\vspace{-3pt}
\noindent{\textbf{Multi-view Reconstruction.}} As shown in \cref{tab:re10k_multiviews} and \cref{tab:crossdata_acid_nview}, our method achieves competitive performance among uncalibrated baselines and remains competitive with pose-free and pose-required methods. When using a similar number of Gaussian primitives as the baselines, our approach achieves strong reconstruction performance. Notably, even when using only 10-28\% of the Gaussian primitives, our method still maintains competitive results. These results suggest that our explicit density control enables a compact yet high-fidelity 3D Gaussian representation, leading to improved efficiency without sacrificing reconstruction quality.

\noindent{\textbf{Two-view Sparse Reconstruction.}} \cref{tab:nvs_2view_acid} demonstrates the novel view synthesis performance on the ACID dataset under the challenging two-view setting. Our method achieves superior performance compared to the uncalibrated baseline. Moreover, it remains competitive with both pose-required and pose-free approaches, including feed-forward 3DGS methods and other baselines, demonstrating robust performance even with only two sparse input views.

\begin{figure*}[t]
    \centering
    \includegraphics[width=1.0\textwidth]{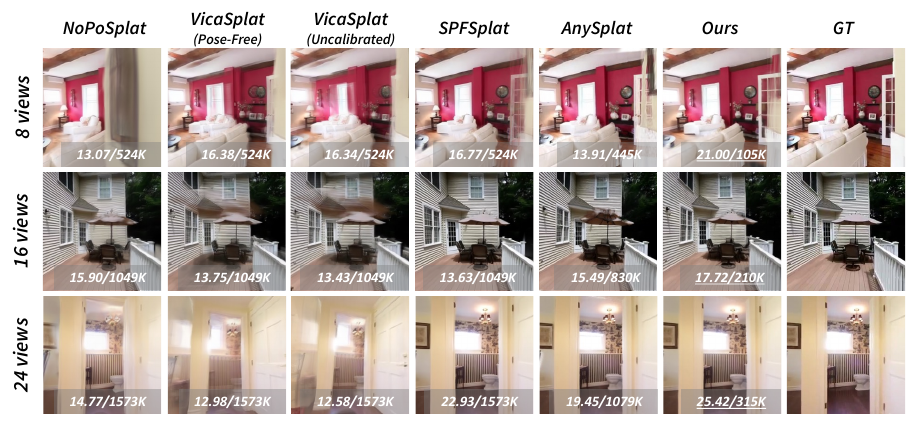}
    \vspace{-10pt}
    \caption{\textbf{Qualitative comparisons of novel view synthesis on RE10K~\cite{zhou2018re10k}.} The bottom-right corner of each image shows the PSNR of the rendered view and the number of Gaussian primitives used for scene reconstruction. 
    Our method achieves high-quality rendering even with substantially fewer primitives, while consistently outperforming competing approaches.}
    \label{fig:qual}
    \vspace{-20pt}
\end{figure*}

\noindent{\textbf{Qualitative Comparison.}} \cref{fig:qual} shows qualitative comparisons on RE10K under different input-view settings. Our method produces sharper structures and more faithful scene details compared to existing approaches. Notably, even with substantially fewer Gaussian primitives (24-29\%), our approach maintains high rendering quality while reducing blurring artifacts, resulting in visually more accurate renderings.

\noindent{\textbf{Generalization Under Larger-Scale Data.}} While \cref{tab:crossdata_acid_nview} evaluates cross-dataset generalization
from RE10K to ACID, we further examine whether the advantage of \OURMODEL~
persists under a larger-scale training setting. Specifically, we train on
RE10K+DL3DV and evaluate zero-shot performance on DTU under the two-view setting, a held-out 3D
reconstruction benchmark. As shown in Tab.~\ref{tab:ood_performance},
F$^{4}$Splat outperforms NoPoSplat across all metrics, indicating that our
predictive densification strategy remains effective when scaling the training
data.
\begin{table}[!t]
\centering
\setlength{\tabcolsep}{0.08cm}
\renewcommand{\arraystretch}{0.9}

\begin{minipage}[t]{0.49\linewidth}
\caption{\textbf{Novel view synthesis performance comparison under 2-view setting.}
Results on ACID~\cite{liu2021acid} dataset.}
\label{tab:nvs_2view_acid}
\vspace{-5pt}

\resizebox{\linewidth}{!}{
\begin{tabular}{llcccc}
\toprule
& \multirow{2}{*}{\textbf{Method}}
& \multicolumn{4}{c}{\textbf{Average}} \\
\cmidrule(lr){3-6}
& 
& \#GS$\downarrow$ & LPIPS$\downarrow$ & SSIM$\uparrow$ & PSNR$\uparrow$ \\
\midrule

\multirow{3}{*}{\shortstack[l]{\emph{Pose-} \\ \emph{Required}}}
& pixelNeRF
& - & 0.533 & 0.561 & 20.32 \\

& pixelSplat 
& \underline{393K} & 0.195 & 0.779 & 25.82 \\

& MVSplat 
& \underline{131K} & 0.196 & 0.773 & 25.51 \\

\midrule

\multirow{6}{*}{\shortstack[l]{\emph{Pose-Free}}}
& DUSt3R
& - & 0.447 & 0.411 & 16.29 \\

& MASt3R
& - &  0.461 & 0.409 &16.18 \\

& CoPoNeRF
& - & 0.406 & 0.606 & 20.95 \\

& NoPoSplat 
& \underline{131K} & 0.189 & 0.781 & 25.96 \\

& VicaSplat
& \underline{131K} & 0.201 & 0.757 & 25.44 \\

& SPFSplat
& \underline{131K} & 0.176 & 0.807 & 26.80 \\
\midrule

\multirow{3}{*}{\shortstack[l]{\emph{Uncalibrated}}}
& VicaSplat 
& \underline{131K} & 0.218 & 0.726 & 24.55 \\
& \textbf{\OURMODEL}$_{\tau^+}$
& \textbf{26K} & \underline{0.201} & \underline{0.772} & \underline{25.75} \\

& \textbf{\OURMODEL}$_{\tau^-}$
& \underline{131K} & \textbf{0.178} & \textbf{0.794} & \textbf{26.26} \\

\bottomrule
\end{tabular}
}
\end{minipage}
\hspace{0.02\linewidth}
\begin{minipage}[t]{0.47\linewidth}
    \captionsetup{type=table}
    \caption{\textbf{Ablation Studies.} Experiments are conducted with 24 input views, where the Gaussian budget is fixed to 20\% of the maximum number of Gaussians. (a)-(b) compare different Gaussian allocation strategies, (c) removes the level-wise Gaussian supervision during training, and (d) removes the scene-scale regularization term.}
    \label{tab:ablation}
    \vspace{-5pt}
    \centering
    \setlength{\tabcolsep}{0.12cm}
    \renewcommand{\arraystretch}{0.95}
    \resizebox{\linewidth}{!}{
    \begin{tabular}{clccc}
        \toprule
        & \textbf{Variant} & \textbf{LPIPS}$\downarrow$ & \textbf{SSIM}$\uparrow$ & \textbf{PSNR}$\uparrow$ \\
        \midrule
        (a) & Rand. based allocation            & 0.194 & 0.828 & 24.68 \\
        (b) & Freq. based allocation   & 0.160 & 0.841 & 25.36 \\
        (c) & w/o level-wise GS train            & 0.192 & 0.813 & 24.25 \\
        (d) & w/o scene scale reg.        & 0.712 & 0.006 & \phantom{0}4.82 \\
        (e) & \textbf{Ours}               & \textbf{0.143} & \textbf{0.854} & \textbf{25.47} \\
        \bottomrule
    \end{tabular}
    }
\end{minipage}
\vspace{-15pt}
\end{table}
\subsection{Ablation Studies and Analysis}
\noindent{\textbf{Densification-Score-Guided Allocation.}}
We validate the effectiveness of our learned densification score for Gaussian allocation by comparing it with two alternative strategies, (a) and (b) in \cref{tab:ablation}.
For (a) \textit{random-based allocation}, we replace the learned densification score with a uniform score, which leads to random Gaussian allocation.
For (b) \textit{frequency-based allocation}, we compute a heuristic frequency score by resizing the input image to each scale level and applying a Sobel filter to emphasize high-frequency (edge) regions.
Both alternatives underperform, demonstrating that the learned densification score provides a more effective signal for allocating Gaussians than either random selection or simple frequency-based heuristics. 
Qualitatively, \cref{fig:densification_score_analysis} further illustrates the behavior of the learned densification score.
The predicted maps of the top example assign higher scores to structurally and visually complex regions (e.g., object boundaries, textured areas, and fine details), indicating where increased Gaussian density is likely to be beneficial.
Moreover, as in the example below, overlapping areas observed from multiple context images tend to have lower scores, reflecting redundancy-awareness across views.
This encourages allocating Gaussians to regions that need more capacity while avoiding redundancy in well-covered areas, improving efficiency and fidelity under a fixed budget.

\noindent{\textbf{Level-Wise Gaussian Supervision.}}
Next, we examine the impact of level-wise Gaussian supervision during training. If we supervise only the final Gaussian set $\mathcal{G}_\tau$, the sampled threshold $\tau$ causes some scale-level Gaussians to be excluded from training in each iteration, resulting in less stable optimization. As shown in \cref{tab:ablation}, removing level-wise supervision clearly degrades performance (c), confirming the benefit of supervising multi-scale Gaussians during training.

\noindent{\textbf{Scene-Scale Regularization.}}
Finally, we ablate the scene-scale regularization loss in~\cref{tab:ablation} (d). Without this term, training becomes highly unstable, and the model fails to learn a meaningful reconstruction, leading to training failure. In the uncalibrated setting, where the model must jointly predict camera parameters and scene geometry, this simple regularizer is crucial for stable optimization.

\noindent{\textbf{VGGT-prior training strategy.}}
\cref{tab:vggt_prior_training} analyzes how to effectively leverage the VGGT
prior for feed-forward 3DGS training. The AnySplat-like alternative performs
input-view training with frozen VGGT distillation, whereas our strategy enables
stable novel-view training through aligned target-camera supervision and
scene-scale regularization. As shown in \cref{tab:vggt_prior_training}, our
strategy achieves better reconstruction quality while reducing training time.
\begin{table}[t]
\centering
\begin{minipage}[t]{0.44\linewidth}
\centering
\caption{\textbf{Generalization on DTU under larger-scale training.}
Models are trained on RE10K~\cite{zhou2018re10k} and DL3DV~\cite{ling2024dl3dv}, and evaluated on the DTU~\cite{jensen2014dtu} dataset.
}
\vspace{-10pt}
\label{tab:ood_performance}
\resizebox{\linewidth}{!}{
\begin{tabular}{lcccc}
\toprule
Method
& \#GS$\downarrow$ & PSNR$\uparrow$ & SSIM$\uparrow$ & LPIPS$\downarrow$ \\
\midrule
 NoPoSplat & 131K & 17.51 & 0.570 & 0.304 \\
\OURMODEL & \textbf{26K} & \textbf{19.76} & \textbf{0.647} & \textbf{0.271} \\
\bottomrule
\end{tabular}
}
\end{minipage}
\hfill
\begin{minipage}[t]{0.53\linewidth}
\centering
\caption{\textbf{Effect of our VGGT-prior training strategy.} Our strategy leverages the VGGT prior more stably than the AnySplat-like alternative, enabling stable novel-view training while reducing training time.}
\vspace{-10pt}
\label{tab:vggt_prior_training}
\resizebox{\linewidth}{!}{
\begin{tabular}{lcccc}
\toprule
Strategy & LPIPS$\downarrow$ & SSIM$\uparrow$ & PSNR$\uparrow$ & Train. Time$\downarrow$ \\
\midrule
AnySplat-like & 0.244 & 0.711 & 20.59 & 5H 35M \\
Ours & \textbf{0.171} & \textbf{0.829} & \textbf{24.62} & \textbf{3H 03M} \\
\bottomrule
\end{tabular}
}
\end{minipage}
\vspace{-15pt}
\end{table}

\vspace{-10pt}
\section{Conclusion}
\vspace{-10pt}
We present \OURMODEL, a feed-forward 3DGS framework that reconstructs compact representations from sparse, uncalibrated inputs. By introducing a feed-forward densification through a densification-score-guided allocation strategy, our method adaptively distributes Gaussians according to spatial complexity and multi-view overlap, enabling explicit control over the final Gaussian budget without retraining. By allocating more primitives to informative regions while avoiding redundancy in simple or overlapping areas, \OURMODEL\ produces compact Gaussian representations while achieving high reconstruction fidelity. Experiments demonstrate that our approach achieves competitive or superior novel-view synthesis performance compared to prior feed-forward methods while requiring significantly fewer Gaussians, highlighting the effectiveness of spatially adaptive Gaussian allocation for efficient feed-forward 3DGS reconstruction.
\vspace{8pt}

\noindent{\textbf{Acknowledgments.}}
This work was supported by Institute of Information \& Communications Technology Planning \& Evaluation (IITP) grant funded by the Korea government (MSIT) (No. RS-2024-00443251, Accurate and Safe Multimodal, Multilingual Personalized AI Tutors, 30\%), the National Research Foundation of Korea (NRF) grant funded by the Korea government (MSIT) (NRF-2023R1A2C2005373, 30\%), the Institute of Information \& communications Technology Planning \& Evaluation (IITP) grant funded by the Korean government (MSIT) (No. RS-2024-00457882, AI Research Lab Project, 30\%), and Institute of Information \& communications Technology Planning \& Evaluation (IITP) grant funded by the Korean government (MSIT) (RS-2025-25442149, LG AI STAR Talent Development Program for Leading Large-Scale Generative AI Models in the Physical AI Domain, 10\%).

\appendix
\setcounter{section}{0}
\renewcommand{\thesection}{S\arabic{section}}

\begin{center}
    {\LARGE \bfseries Supplementary Material \par}
    \vspace{0.5em}
    {\large \bfseries F$^4$Splat: Feed-Forward Predictive Densification for Feed-Forward 3D Gaussian Splatting\par}
\end{center}

\vspace{1em}

\setcounter{page}{1}
\setcounter{table}{0}
\renewcommand{\thetable}{S\arabic{table}}
\setcounter{figure}{0}
\renewcommand{\thefigure}{S\arabic{figure}}

\section{Budget Matching Algorithm}
To efficiently determine the threshold $\bar{\tau}$ that matches a target Gaussian budget $\bar{N}_\mathcal{G}$, we precompute a threshold--budget lookup table, $\tilde{\boldsymbol{\tau}}$ and $\tilde{\mathbf{N}}_\mathcal{G}$, from the predicted multi-level densification score maps $\{\{\hat{\mathbf{D}}_i^l\}_{i=1}^{N_\tec}\}_{l=1}^{L-1}$ using \cref{alg:threshold_budget_lookup}.

\begin{algorithm}[ht]
\caption{\textbf{Precomputing the threshold--budget lookup table}}
\label{alg:threshold_budget_lookup}
\begin{algorithmic}[1]
\Require Multi-level score maps $\{\{\hat{\mathbf{D}}_i^l\}_{i=1}^{N_\text{ctx}}\}^{L-1}_{l=1}$ for all views and levels
\Ensure A sorted threshold list $\tilde{\boldsymbol{\tau}}$ and the corresponding Gaussian-count delta list $\tilde{\mathbf{N}}_\mathcal{G}$

\State Initialize all $\Delta N_i^l$ to $3$ for all values with the same resolution as $\hat{\mathbf{D}}_i^l$.
\For{$l=L-1,L-2,\dots,2$}
    \State $\{\mathbf{A}_i^l\}_{i=1}^{N_\text{ctx}} \leftarrow \left\{ \mathds{1}_{\{\hat{\mathbf{D}}_i^l \ge \mathrm{Up}(\hat{\mathbf{D}}_i^{l-1};2)\}}\right\}_{i=1}^{N_\text{ctx}}$
    
    \State $\{\Delta N_i^{l-1} \}_{i=1}^{N_\text{ctx}} \leftarrow \left\{ \Delta N_i^{l-1} + \mathrm{SumPool}_{2\times2}\!\left(\Delta N_i^{l} \odot \mathbf{A}_i^l\right)\right\}_{i=1}^{N_\text{ctx}}$
    \State $\{\Delta N_i^{l} \}_{i=1}^{N_\text{ctx}} \leftarrow \left\{ \Delta N_i^{l} \odot (\mathbf{1}-\mathbf{A}_i^l) \right\}_{i=1}^{N_\text{ctx}}$
\EndFor

\State $\boldsymbol{{\tau}} = \mathrm{Concat}\!\left(\left\{\mathrm{Flatten}(\hat{\mathbf{D}}_i^l)\right\}_{i=1,\dots,N_{\tec};\,l=1,\dots,L-1}\right)$
\State $\boldsymbol{\Delta}\mathbf{N}_\mathcal{G} = \mathrm{Concat}\!\left(\left\{\mathrm{Flatten}(\Delta N_i^l)\right\}_{i=1,\dots,N_{\tec};\,l=1,\dots,L-1}\right)$

\State $\boldsymbol{\pi} = \text{argsort}(\boldsymbol{{\tau}})$
\Comment{descending sorting permutation}

\State $\tilde{\boldsymbol{{\tau}}} = \boldsymbol{{\tau}}[\boldsymbol{\pi}]$

\State ${\boldsymbol{\Delta} \mathbf{N}}_\mathcal{G} = \boldsymbol{\Delta}\mathbf{N}_\mathcal{G}[\boldsymbol{\pi}]$

\State $\tilde{\mathbf{N}}_\mathcal{G} =  N_{\text{ctx}}H^1W^1 \cdot \mathbf{1} + \text{Cumsum}\left( {\boldsymbol{\Delta} \mathbf{N}}_\mathcal{G} \right) $
\Comment{sorted in ascending order, unlike $\tilde{\boldsymbol{\tau}}$ }

\State \Return $\tilde{\boldsymbol{{\tau}}}, \tilde{\mathbf{N}}_\mathcal{G}$
\end{algorithmic}
\end{algorithm}

Let $\tilde{\mathbf{N}}_\mathcal{G} = \big([\tilde{\mathbf{N}}_\mathcal{G}]_1,\dots,[\tilde{\mathbf{N}}_\mathcal{G}]_K\big)$
denote the Gaussian-count lookup table sorted according to the threshold list
$\tilde{\boldsymbol{\tau}}$. Under the assumption that all densification score values are unique,
the difference between two adjacent entries is upper-bounded by $4^{L-1}-1$:
\[
[\tilde{\mathbf{N}}_\mathcal{G}]_{k+1} - [\tilde{\mathbf{N}}_\mathcal{G}]_{k} \le 4^{L-1}-1.
\]
This follows because lowering the threshold activates exactly one new score value, which changes the allocation at only one spatial position. The largest possible increase occurs when the activated position corresponds to the coarsest valid level and represents all unresolved descendants across finer levels. Since each finest-level position contributes $3$ Gaussians, the maximum increment becomes
\[
3\sum_{j=0}^{L-2}4^j = 4^{L-1}-1.
\]
Therefore, for any target budget $\bar{N}_\mathcal{G}$, there always exists an index $k^\star$ such that
\[
[\tilde{\mathbf{N}}_{\mathcal{G}}]_{k^\star} \le \bar{N}_\mathcal{G}
< [\tilde{\mathbf{N}}_{\mathcal{G}}]_{k^\star} + (4^{L-1}-1),
\]
which implies that the threshold $\bar{\tau}=[\tilde{\boldsymbol{\tau}}]_{k^\star}$ satisfies Eq.~(4) of the main paper.
Ultimately, we can efficiently obtain the threshold $\bar{\tau}$ via \cref{alg:find_budget_threshold}.
\begin{algorithm}[ht]
\caption{\textbf{Finding the budget-matching threshold}}
\label{alg:find_budget_threshold}
\begin{algorithmic}[1]
\Require $\tilde{\boldsymbol{\tau}}$, $\tilde{\mathbf{N}}_{\mathcal{G}}$, $\bar{N}_{\mathcal{G}}$
\Ensure $\bar{\tau}$ satisfying the target budget $\bar{N}_{\mathcal{G}}$

\State $k^\star \leftarrow \max \left\{ k \mid [\tilde{\mathbf{N}}_{\mathcal{G}}]_{k} \le \bar{N}_{\mathcal{G}} \right\}$ 
\Comment{found by binary search in $O(\log |\tilde{\mathbf{N}}_{\mathcal{G}}|)$ time}

\State $\bar{\tau} \leftarrow [\tilde{\boldsymbol{\tau}}]_{k^\star}$
\State \Return $\bar{\tau}$
\end{algorithmic}
\end{algorithm}
\begin{table}[t]
\centering
\begin{minipage}[t]{0.51\linewidth}
    \captionsetup{type=table}
    \caption{\textbf{Camera pose estimation accuracy on RE10K and ACID.} Following prior works~\cite{sarlin2020superglue,edstedt2024roma}, we quantify accuracy via the AUC of the cumulative rotation error curve, using rotation thresholds of $5^\circ$, $10^\circ$, and $20^\circ$.
    }
    \label{sup_tab:camera_pose}
    \centering
    \setlength{\tabcolsep}{0.12cm}
    \renewcommand{\arraystretch}{0.95}
    \resizebox{\linewidth}{!}{
    \begin{tabular}{lccc ccc}
        \toprule
        & \multicolumn{3}{c}{\textbf{RE10K}} 
        & \multicolumn{3}{c}{\textbf{ACID}} \\
        \cmidrule(lr){2-4} \cmidrule(lr){5-7} 
        \textbf{Method} 
        & 5$^\circ$~$\uparrow$ & 10$^\circ$~$\uparrow$ & 20$^\circ$~$\uparrow$
        & 5$^\circ$~$\uparrow$ & 10$^\circ$~$\uparrow$ & 20$^\circ$~$\uparrow$ \\
        \midrule

        DUSt3R 
        & 0.301 & 0.495 & 0.657 
        & 0.166 & 0.304 & 0.437 \\

        MASt3R 
        & 0.372 & 0.561 & 0.709 
        & 0.234 & 0.396 & 0.541 \\

        VGGT
        & \underline{0.335} & \underline{0.531} & \underline{0.696} 
        & 0.219 & 0.399 & 0.576 \\

        AnySplat (RE10K)
        & 0.315 & 0.522 & 0.694 
        & \textbf{0.265} & \underline{0.445} & \underline{0.608} \\

        \textbf{F$^\mathbf{4}$Splat} (RE10K)
        & \textbf{0.541} & \textbf{0.704} & \textbf{0.814} 
        & \underline{0.262} & \textbf{0.449} & \textbf{0.618} \\
        \bottomrule
    \end{tabular}
    }
\end{minipage}
\hfill
\begin{minipage}[t]{0.44\linewidth}
    \captionsetup{type=table}
    \caption{\textbf{Computational overhead of Spatially Adaptive Gaussian Allocation.}
    We compare the uniform pixel-to-Gaussian allocation with our adaptive allocation. 
    For adaptive allocation, the total number of Gaussians is reduced by 20\% compared to the uniform allocation.}
    \vspace{-4pt}
    \label{sup_tab:adaptive_cost}
    \centering
    \setlength{\tabcolsep}{0.12cm}
    \renewcommand{\arraystretch}{0.95}
    \resizebox{\linewidth}{!}{
\begin{tabular}{lcc}
\toprule
{\textbf{Allocation}} & {\textbf{Peak allocated}} & {\textbf{Inference}}\\ 
\textbf{strategy} & \textbf{VRAM (GB) } & \textbf{time (s)} \\
\midrule
Uniform  & 8.699 & 0.440 \\
Adaptive & 8.855 (+1.8\%) & 0.488 (+10.1\%) \\
\bottomrule
\end{tabular}
    }
\end{minipage}
\end{table}

\section{Additional Experiments}

\noindent{\textbf{Relative Pose Estimation.}}
Following prior work~\cite{sarlin2020superglue, edstedt2024roma}, 
we additionally evaluate the quality of the predicted camera poses using the relative rotation accuracy metric. Specifically, we compare DUSt3R~\cite{wang2024dust3r}, MASt3R~\cite{leroy2024mast3r}, and VGGT~\cite{wang2025vggt} against AnySplat and \OURMODEL, which are trained with multi-view supervision on RE10K.
We construct the evaluation sets on RE10K and ACID following the same protocol as NoPoSplat~\cite{ye2024nopo}, but, unlike NoPoSplat, we directly use the camera parameters predicted by our model without any test-time optimization. 
As shown in \cref{sup_tab:camera_pose}, our model, trained on the RE10K dataset, achieves higher accuracy across all thresholds. 
Furthermore, despite not being trained on the ACID dataset~\cite{liu2021acid}, \OURMODEL\, generalizes effectively to this unseen dataset, achieving the best performance overall while outperforming all baselines at $10\degree$ and $20\degree$ and remaining competitive at $5\degree$. 
These results indicate that, although our geometry backbone is initialized from VGGT, the proposed architecture and training scheme further refine and extend its geometric reasoning, leading to consistently stronger pose estimation performance.

\noindent{\textbf{Computational cost of Spatially Adaptive Gaussian Allocation.}}
\cref{sup_tab:adaptive_cost} reports the computational overhead of introducing Spatially Adaptive Gaussian Allocation. The peak allocated VRAM increases by only $1.8\%$, and the inference time increases by $10.1\%$. As discussed in the \textit{Multi-Scale Prediction} paragraph of Sec. 3.2 in the main paper, the proposed design predicts multi-scale maps efficiently, making adaptive Gaussian allocation possible with minimal extra cost.

\section{Additional Details}
For all experiments, we freeze only the image tokenizer, DINOv2~\cite{oquab2023dinov2}, while fine-tuning all other components, including the frame-wise and global attention layers, camera and register tokens, and the point and camera prediction heads. We use a base learning rate of $2\!\times\!10^{-4}$ for all experiments, and apply a learning rate scaled by $1/10$ to the fine-tuned components inherited from the pretrained model. For evaluation on RE10K~\cite{zhou2018re10k} and ACID~\cite{liu2021acid}, we follow NoPoSplat and apply test-time camera pose optimization. The initial target camera parameters are obtained using the target-aligned projection formulation described in Sec.~3.3 of the main paper. For the ablation studies, we adopt a 5$\times$ reduced-iteration training schedule, where all iteration-related hyperparameters are reduced by a factor of 5. For the AnySplat results in Fig.~1 (a), we control the number of Gaussians by adjusting the voxel size. As the voxel size increases, we proportionally enlarge the scale of the output Gaussians so that their spatial coverage increases accordingly. 

\begin{figure*}[!t]
     \centering
     \includegraphics[trim=0 0 0 0,clip, width=1.0\linewidth]{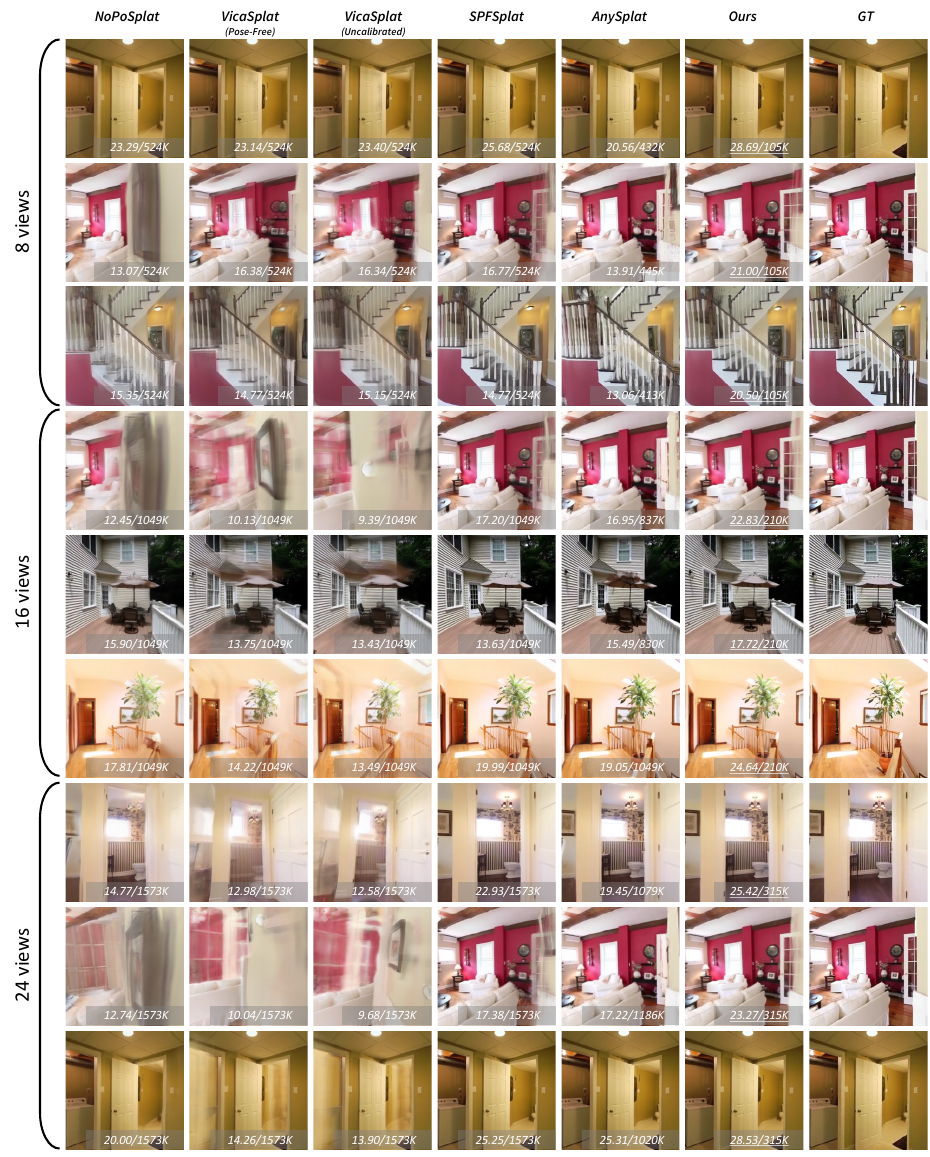}
     \caption{\textbf{Additional qualitative comparisons of novel view synthesis on RE10K~\cite{zhou2018re10k} dataset.}}
     \label{fig:supple_qual_re10k}
\end{figure*}


\clearpage

\bibliographystyle{splncs04}
\bibliography{main}
\end{document}